%% file: main.tex
\documentclass[10pt,twocolumn,letterpaper]{article}
\pdfoutput=1
\usepackage{cvpr}
\usepackage{times}
\usepackage{epsfig}
\usepackage{graphicx}
\usepackage{amsmath}
\usepackage{amssymb}
\usepackage{booktabs}

\usepackage[dvipsnames]{xcolor}

\definecolor{sadness}{RGB}{74, 65, 42}

\usepackage[pagebackref=true,breaklinks=true,letterpaper=true,colorlinks,bookmarks=false]{hyperref}

\cvprfinalcopy 

\def\Mnasfpn{MnasFPN~}
\def\Mnasfpnnospace{MnasFPN}

\ifcvprfinal\fi

\begin{document}

\title{\Mnasfpn: Learning Latency-aware Pyramid Architecture \\ for Object Detection on Mobile Devices}

\author{
	\small
	\begin{tabular}{c c c c c c c }                                                          
        \bf Bo Chen &
		\bf Golnaz Ghiasi &
		\bf Hanxiao Liu &
		\bf Tsung-Yi Lin &
		\bf Dmitry Kalenichenko &
		\bf Hartwig Adam &
		\bf Quoc V. Le \\                                        
		\multicolumn{7}{c}{Google Research} \\       
		\multicolumn{7}{c}{bochen.caltech@gmail.com,\{golnazg, hanxiaol, tsungyi, dkalenichenko, hadam, qvl\}@google.com} \\
	\end{tabular}                                                                       
}

\maketitle

\begin{abstract}
Despite the blooming success of architecture search for vision tasks in resource-constrained environments, the design of on-device object detection architectures have mostly been manual. The few automated search efforts are either centered around non-mobile-friendly search spaces or not guided by on-device latency. We propose \Mnasfpnnospace, a mobile-friendly search space for the detection head, and combine it with latency-aware architecture search to produce efficient object detection models. The learned \Mnasfpn head, when paired with MobileNetV2 body, outperforms MobileNetV3+SSDLite by $1.8$ mAP at similar latency on Pixel. It is both $1$ mAP more accurate and $10\%$ faster than NAS-FPNLite. Ablation studies show that the majority of the performance gain comes from innovations in the search space. Further explorations reveal an interesting coupling between the search space design and the search algorithm, for which the complexity of \Mnasfpn search space is opportune\footnote{Implementation is available at Tensorflow Object Detection API \url{https://github.com/tensorflow/models/tree/master/research/object_detection}}.
\end{abstract}

\input{introduction.tex}

\input{related_work.tex}

\input{methods.tex}

\input{experiments.tex}

\input{conclusions.tex}

{\small
\bibliographystyle{ieee_fullname}
\bibliography{references}
}
\appendix

\input{appendix.tex}

\cleardoublepage
\end{document}

%% file: introduction.tex
\section{Introduction}
\label{sec:intro}

\begin{table}[ht]
\centering
\resizebox{\linewidth}{!}{
\begin{tabular}{@{}lcccc@{}}
\toprule
\textbf{Model} & \textbf{Test-dev mAP} & \textbf{Latency} & \textbf{MAdds} & \textbf{Params} \\ \midrule
MobileNetV3$^\dagger$ + SSDLite & $22.0$~\cite{howard2019searching} & $119^*$ & $0.51$B & $3.22$M \\
MobileNetV3 + SSDLite & $22.0$~\cite{howard2019searching} & $137^*$ & $0.62$B & $4.97$M \\
MobileNetV2 + SSDLite & $22.1$~\cite{sandler2018mobilenetv2} & $163^*$ & $0.8$B & $4.3$M \\
MnasNet-A1 + SSDLite & $23.0$~\cite{tan2019mnasnet} & $174^*$ & $0.8$B & $4.9$M \\
MobileNetV2 + NAS-FPNLite & $25.1$~\cite{ghiasi2019fpn} & $202^*$ & $0.98$B & $2.02$M \\
\midrule
MobileNetV2 + \Mnasfpn $\ddagger$ & $23.8$ & $121$ & $0.53$B & $1.29$M \\
MobileNetV3 + \Mnasfpn & $25.5$ & $168$ & $0.77$B & $3.46$M \\ 
MobileNetV2 + \Mnasfpn & $26.1$ & $183$ & $0.92$B & $2.50$M \\ \bottomrule
\end{tabular}
}
\vspace{0.2pt}
\caption{\Mnasfpn variations compared with other mobile detection models on COCO {\it test-dev}. Latency numbers with `*` are re-measured in the same configuration (same benchmarker binary and same device) as \Mnasfpn models to ensure fairness of comparison. Models with $\dagger$ employs the channel-halving trick~\cite{howard2019searching}. Models with $\ddagger$ was obtained with a depth multiplier of $0.7$ on both head and backbone.}
\label{tab:testdev}
\end{table}

Designing neural network architectures for efficient deployment on mobile devices is not an easy task: one has to judiciously trade off the amount of computation with accuracy, while taking into consideration the set of operations that are supported and favored by the devices. Neural architecture search (NAS, \cite{zoph2016neural}) provides the framework to automate the design process, where a RL controller will learn to generate fast and accuracy models within a user-specified search space. While the focus of NAS papers have been on improving the search algorithm, the search space design remains a critical performance factor that is less visited.

Despite the significant advances on NAS for image classification both in the server setting \cite{zoph2016neural, tan2019efficientnet} and in the mobile setting~\cite{tan2019mnasnet, cai2018proxylessnas, howard2019searching, wu2019fbnet, dai2019chamnet}, relatively fewer attempts~\cite{ghiasi2019fpn, chen2019detnas, wang2019nasfcos} focus on object detection. This is in part because the additional complexity in the search space of the detection head relative to the backbone. 
The backbone is a feature extractor that sequentially extracts features at increasingly finer scales, which behaves the same way as the feature extractor for image classification. Therefore, current NAS approaches either repurpose classification feature extractors for detection~\cite{howard2019searching, tan2019mnasnet, tan2019efficientnet}, or search the backbone while fixing the detection head~\cite{chen2019detnas}.
Since the backbone is composed of a sequence of layers, its search space is sequential. In contrast, a detection head could be highly non-sequential. It needs to fuse and regenerate features across multiple scales for better class prediction and localization. The search space therefore includes what features to fuse, as well as how often and in what order to fuse them. This is a challenging task that few NAS frameworks have demonstrated the ability to handle.

One exception is NAS-FPN~\cite{ghiasi2019fpn}, which was the first NAS paper that tackles the non-sequential search space of the detection head. It demonstrates state-of-the-art performance when optimized for accuracy only, and its manually designed variant called NAS-FPNLite performs competitively on mobile devices. However, NAS-FPNLite is limited in three aspects. 1) The search process that produces the architecture is not guided by computational complexity or on-device latency; 2) The architecture was manually adapted to work with mobile devices, of which the process may be further optimized; 3) The original NAS-FPN search space was not tailored towards mobile use cases. 

Our work addresses the above limitations. We propose a search space called \Mnasfpnnospace, which is specifically designed for mobile devices where depthwise convolutions are reasonably optimized. Our search space re-introduces the inverted residual block~\cite{sandler2018mobilenetv2}, which is proven to be effective for mobile CPUs, into the detection head. We conduct NAS on the search space that is guided by on-device latency signals. The search found an architecture that is remarkably simple yet highly performant.

Our contributions include:
1) A {\bf mobile-specific search space} for the detection head;
2) The {\bf first attempt} to conduct latency-aware search for object detection;
3) A set of {\bf detection head architectures} that outperform SSDLite~\cite{sandler2018mobilenetv2} and NAS-FPNLite~\cite{ghiasi2019fpn};
4) {\bf Ablation studies} showing that our search space design is judiciously chosen for the current NAS controller.

%% file: related_work.tex
\section{Related Work}
\subsection{Mobile Object Detection Models}
The most common detection models on mobile devices are manually designed by experts. Among them are single-shot detectors such as YOLO~\cite{redmon2016you}, SqueezeDet~\cite{wu2017squeezedet}, and Pelee~\cite{wang2018pelee} as well as two-stage detectors, such as Faster RCNN~\cite{ren2015faster}, R-FCN~\cite{dai2016r}, and ThunderNet~\cite{qin2019thundernet}.

SSDLite~\cite{sandler2018mobilenetv2} is the most popular light-weight detection head architecture. It replaces the expensive $3\times3$ full convolutions in the SSD head~\cite{liu2016ssd} with separable convolutions to reduce computational burden on mobile devices. This technique is also employed by NAS-FPNLite~\cite{ghiasi2019fpn} to adapt NAS-FPN to mobile devices. SSDLite and NAS-FPNLite are paired with efficient backbones such as MobileNetV3~\cite{howard2019searching} to produce state-of-the-art mobile detectors. Since we design mobile-friendly detection heads, both SSDLite and NAS-FPNLite are crucial baselines to showcase our effectiveness.

\subsection{Architecture Search for Mobile Models}
Our NAS search is guided by latency signals that come from on-device measurements. Latency-aware NAS was first popularized by NetAdapt~\cite{yang2018netadapt} and AMC~\cite{he2018amc} to learn channel sizes for a pre-trained model. A look-up table (LUT) was used to efficiently estimate the end-to-end latency of a network based on the latency sum of its parts. This idea was then extended in MnasNet~\cite{tan2019mnasnet} to search for generic architecture parameters using the NAS framework~\cite{zoph2016neural}, where a RL controller learns to generate efficient architectures after observing the latency and accuracy of thousands of architectures. This framework was successfully adopted by MobileNetV3~\cite{howard2019searching} to produce the current state-of-the-art architectures for mobile CPU.  

The MnasNet-style search was not accessible to researchers with limited resources. Therefore a large body of the NAS literature~\cite{cai2018proxylessnas, wu2019fbnet,cai2019once} focus on improving the search efficiency. These methods capitalize on the idea of hyper-network and weight-sharing~\cite{cai2018proxylessnas, bender2018understanding,pham2018efficient} to boost search efficiency. Despite the success in mobile classification, these efficient search techniques have not been extended to highly non-sequential search spaces in resource-constrained cases, hence have not seen many applications in mobile object detection.

\subsection{Architecture Search for Object Detection}

Due to the above-mentioned non-sequential nature of search in object detection, NAS work on object detection has generally been limited.

NAS-FPN~\cite{ghiasi2019fpn} was the pioneering work that tackles detection head search. It proposes an overarching search space based on feature pyramid networks~\cite{lin2017feature}. The design covers many popular detection heads. Our work is primarily inspired by NAS-FPN, but with the goal of innovating a search space that is more mobile-friendly. 

Another pioneering work was Auto-Deeplab~\cite{liu2019auto}, which extended NAS searches to semantic segmentation. Our work faces the similar challenge of learning the connectivity pattern across feature resolutions.

DetNAS~\cite{chen2019detnas} focuses on improving the efficiency of searching for the detection body. It deals with the unmanageable computation caused by the need for ImageNet pre-training for every sampled architecture during search. Our work instead searches for the head only.

More recently, NAS-FCOS~\cite{wang2019nasfcos} extends  weight-sharing to the detection head in order to accelerate the search process for object detection. Similar to NAS-FPN, their search space for the detection head is based on full convolutions and not targeted for mobile. Our work is complementary to theirs, in that our latency-aware search based on a mobile-friendly search space could be accelerated with their weight-sharing search strategy.

On the mobile side, object detection architectures are rarely optimized as a primary target. Rather, they are composed of a light-weight backbone designed for classification and a predefined detection head. A partial list of work that follows this design strategy is ~\cite{howard2019searching, tan2019mnasnet, sandler2018mobilenetv2, zhang2018shufflenet}. 
Our work takes a first step towards directly optimizing object detection architectures for mobile deployment.

%% file: methods.tex
\section{\Mnasfpn}

\begin{figure*}[!t]
    \centering
    \includegraphics[width=1.0\linewidth]{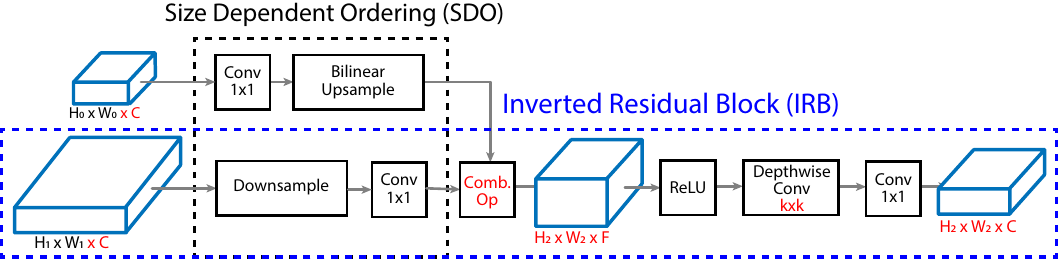}
    \caption{A searchable \Mnasfpn block. \Mnasfpn re-introduces the Inverted Residual Block (IRB) into the NAS-FPN head (Sec.~\ref{sec:irb}). Any path connecting an input and a new feature, as highlighted in blue dashed rectangle, resembles an IRB. \Mnasfpn also employs Size Dependent Ordering (SDO) shown in black rectangle to re-order the resizing operation and the $1\times1$ convolution prior to feature merging (Sec.~\ref{sec:sdo}). Search-able components are highlighted in red (Sec.~\ref{sec:search_space}).}
    \label{fig:search_space}
\end{figure*}

We overload the term \Mnasfpn to mean both our proposed search space and the family of architectures found via NAS, and leave disambiguation to context. Both NAS-FPN(Lite) and \Mnasfpn construct a detection network from a feature extractor backbone and a repeatable cell structure that recursively generates new features by merging pairs of existing features. Each {\it cell} consumes a collection of feature maps at various resolutions, and outputs another collection at the same set of resolutions, thus enabling the structure to be applied repeatedly. A cell is comprised of a collection of {\it blocks}. Each block merges two feature maps at potentially different resolutions into an {\it intermediate feature}, which is processed by a separable convolution and outputted by the block. \Mnasfpn differs from NAS-FPN(Lite) mainly at the block-level, which we describe below.

\subsection{Generalized Inverted Residual Block (IRB)}
\label{sec:irb}

Inverted Residual Blocks (IRBs)~\cite{sandler2018mobilenetv2} are well known block architectures that are widely used in NAS search spaces~\cite{tan2019mnasnet, cai2018proxylessnas, howard2019searching, wu2019fbnet}. The key insight of IRBs is to communicate features in low-dimensions in order to reduce memory impact, and expand the feature dimensions for depthwise convolutions in order to exploit their light-weight nature in mobile CPUs. It has shown superior performance gains over the conventional block design based on separable convolutions. This motivates us to explore the possibility of adopting IRB-like designs in the NAS-FPN search space, where the main challenge and innovation reside in improvising with the non-linear structure in NAS-FPN blocks.


{\bf Expandable intermediate features}: In NAS-FPN, all feature maps share the same, searchable channel size $C$ by design. By comparison, \Mnasfpn gives additional flexibility to the intermediate feature size $F$, which is both searchable and independent from $C$. By adjusting $F$ and $C$, the intermediate feature can serve as either an expansion or a bottleneck. Such network with unequal input and merged feature sizes is an instance of {\it asymmetric FPNs}, as defined in~\cite{kirillov2019panoptic}. A $1\times1$ convolution is applied as needed on each input feature to transform their channel count from $C$ to $F$. 

{\bf Learn-able block count}: In NAS-FPN, the number of blocks in a cell is predetermined. This comes from the feature recycling mechanism where if a block is not consumed by the cell's outputs, its intermediate feature will be added to the output feature with the same resolution and size. In \Mnasfpnnospace, however, the intermediate features often do not have the same channel size $F$ as the output $C$. As a result, unused blocks are frequently discarded, giving additional flexibility in navigating the latency-accuracy trade-off.  

{\bf Cell-wide residuals}: As the connectivity gets thinner, we found that it's helpful to augment the flow of information by adding residuals between every pair of input and output features at the same resolution. 
Similar to IRB, we add {\it ReLU} non-linearity for the intermediate features, but not the output features. This is because the input/output feature channel size $C$ is intended to be small to lessen the burden on memory. Adding lossy non-linearities may unnecessarily throttle the information flow. 

Given the design above, one can traverse a connected path between an input feature and an output feature and see that it resembles an IRB, as shown in Fig.~\ref{fig:search_space}. 

We have not experimented with the MobileNetV3-styled IRB with hard-swish in the search space because their implementations were not optimized at the time of the experiment design for this paper. They are worth re-visiting once efficient kernels for hard-swish become available. We have explored Squeeze-Excite (SE)~\cite{hu2018squeeze}, but much to our surprise, it was not chosen by our NAS controller for top-performing candidates.

\subsection{Size Dependent Ordering (SDO)}
\label{sec:sdo}

Another innovation in \Mnasfpn is the dynamic re-ordering of its reshaping and convolution operations based on the input/output resolutions. We refer to this as Size Dependent Ordering (SDO). More specifically, if the input feature needs to be down-sampled, then down-sampling will happen prior to the $1\times 1$ convolution. On the contrary, if the input feature requires up-sampling, then $1\times 1$ convolution will precede the up-sampling operation. 

This design minimizes compute. For notation simplicity we assume the feature maps are square, and use $R$ to represent both the height and width. When merging feature maps, we need to apply reshaping and $1\times 1$ convolutions when the resolution $R_0$ and channel count $C$ of the input feature do not match the resolution $R$ and channel count $F$ of the intermediate feature. 

If $R_0 > R$ (needs down-sampling), let $R_0 = kR$ where $k\geq2$, and assume down-sampling is performed with $k\times k$ convolution with a stride also equals to $k$, the cost (in MACs) of down-sample-then-$1\times1$ is:
\begin{align}
    Cost_1 =&R \times R \times k \times k \times C + R \times R \times C \times F \label{eq:cost_1}
\end{align}
whereas the cost of $1\times1$-then-down-sample is:
\begin{align}
    Cost_2 =&kR \times kR \times C \times F + R \times R \times k \times k \times F  \label{eq:cost_2}
\end{align}

Assume reasonably that $F\geq 2$, we have $k^2C(F-1)\geq k^2CF/2 \geq CF$, therefore:
\begin{align}
Cost_2 - Cost_1&= R^2\left(k^2C(F-1)+R^2F-CF\right) > 0 \label{eq:proof}
\end{align}
hence proving that the down-sample-then-$1\times 1$ is more economical. The case for $R_0 < R$ (up-sampling) can be proved similarly.


\subsection{\Mnasfpn Search}
\label{sec:search_space}



The feature generation process of \Mnasfpn and all the searchable components are illustrated in Fig.~\ref{fig:search_space}. For each feature generation block, we search for which two input features to merge, the target resolution $R$ and channel count $F$ of the merged feature, the merging operation (addition or SE), and the kernel sizes for the depthwise convolution post merging. For the entire network, we mandate that the input, output and generated features all share the same channel count $C$, which is also searched.

We adopt the architecture search framework in MnasNet~\cite{tan2019mnasnet} to incorporate latency measurements into the search objective. We train an RL controller to propose network architectures to maximize a reward function defined as follows. An architecture $m$ is trained and evaluated on a proxy task. The proxy task is a scaled-down version of the real task, with details in Sec.~\ref{sec:exp_setup}. The proxy task performance, measured in mean average precision $mAP(m)$, as well as the network latency on-device $LAT(m)$ are combined into the following reward function:

\begin{align}
    Reward(m) &= \,\,mAP(m) \times LAT(m)^w
    \label{eq:reward}
\end{align}
where $w < 0$ controls the tradeoff point between latency and accuracy. In theory, $w$ is the slope of the tangent line that cuts the performance trade-off curve at the desired latency. In practice, we observe that architectures around the desired latency will also be optimized, and the performance frontier of our search spaces have similar curvatures, suggesting that $w$ needs to be set only once.

The controller repeatedly proposes candidate architecture $m$, and trains itself based on reward feedback $Reward(m)$ using Proximal Policy Optimization~\cite{schulman2017proximal}. After every search experiment, all the  architectures sampled by the controller trace a performance frontier, as shown in Fig.~\ref{fig:frontiers}. We can then deploy promising architectures along the frontier to the real task.

{\bf Connectivity-based LUT}: We apply detection-specific adaptations to the latency look-up table~\cite{yang2018netadapt, tan2019mnasnet} to estimate $LAT(m)$. Existing LUT approaches do not work for \Mnasfpn because the number of blocks and the connectivity pattern of the head is dynamic. Instead, we compute layer connectivity for each model at run-time to determine the layers to be included in the look-up. The connectivity-based LUT gives high fidelity with on-device measurements ($R^2>0.97$).

\subsection{Connectivity Search}
\label{sec:conn}
Our design of the \Mnasfpn search space is deliberately compact. This is in consideration of the fact that current architecture search algorithms are imperfect~\cite{li2019random}, and larger search spaces {\it do not} always lead to better models. Therefore, search space design is as much about what to include as it is about what {\it not} to include. 

One design we do not include is the search for more general connectivity patterns. It overburdens the MNAS controller but remains valuable as search algorithms continue to improve. Recent work on randomly-wired networks~\cite{xie2019exploring} suggests that search quality may be hampered by design biases in network connectivity in addition to search efficacy. We therefore challenge the connection rule in NAS-FPN where only two features are chosen to be merged each time. Instead, we design a new search space {\bf Conn-Search} that allows merging between $2$ to $D\geq 2$ distinct feature maps with addition ($D=4$ in our experiments). 


%% file: experiments.tex
\section{Experiments}
\label{sec:exp}

We present experimental results to showcase the effectiveness of the proposed \Mnasfpn search space. We report results on COCO object detection. We also added ablation studies to isolate the effectiveness of every component of the search space design as well as latency-aware search.

\subsection{Search Experiments and Models}

\begin{table*}[ht]
\centering
\vspace{0pt}
\scalebox{0.8}{
\begin{tabular}{c|c|c|c|c|c|c}
\toprule[0.2em]
Search spaces & Kernel sizes &  Filter sizes $C$ & Expansion sizes $F$ & SDO & Maximum in-degree & Cardinality\\
\toprule[0.2em]
NAS-FPNLite-S & $3$	& $64$ & - & N & $2$ & $2\times 10^{22}$\\	
No-Expand & $\{3,5,7\}$ & $\{16, 32, 48, 64, 80, 96\}$ & - & Y & $2$ & $2.4\times 10^{27}$ \\
\Mnasfpn & $\{3,5,7\}$ & $\{16, 32, 48, 64, 80, 96\}$ & $\{16, 32, 64, 96, 128, 256, 512\}$ & Y & $2$ & $10^{31}$\\
Conn-Search & $\{3,5,7\}$ & $\{16, 32,48, 64, 80, 96\}$ & $\{16, 32, 64, 96, 128, 256, 512\}$ & Y & $4$ & $3 \times 10^{42}$  \\
\toprule[0.2em]
\end{tabular}
}
\vspace{0.2pt}
\caption{Search space comparisons. The common search parameters (e.g. merge operations,  feature resolutions etc.) are omitted. 
}
\label{table:all-search-spaces}
\end{table*}

We include the following experiments / models. All search spaces allow $5$ internal blocks per cell. 

{\bf \Mnasfpnnospace}: Our proposed search space with searchable \Mnasfpn blocks described in Fig.~\ref{fig:search_space}. 

{\bf NAS-FPNLite~\cite{ghiasi2019fpn}}: NAS-FPN models that are post-hoc modified to be light-weight, where modification refers to replacing full convolutions in the head with separable-convolutions. These are the only set of models that are {\it not} searched via latency-sensitive NAS (Sec.~\ref{sec:search_space}).

{\bf NAS-FPNLite-S}: Modified NAS-FPN search space where full convolutions are replaced with separable-convolutions. A key distinction from NAS-FPNLite is that the modification is done on the search space, instead of post-hoc on the model.

{\bf No-Exand}: We remove and only remove expansion from the \Mnasfpn search space by enforcing $F=C$ for all intermediate features. This serves as an ablation of the expansion in IRB. It differs from NAS-FPNLite-S in that it still retains all other \Mnasfpn designs such as SDO and cell-wide residual, as well as search-able options.

{\bf Conn-Search}: We enlarges the \Mnasfpn search space by allowing between $2$ to $D\geq 2$ distinct inputs per block. Merge operation is limited to addition only. 


A detailed comparison of all the search spaces in the ablation studies are listed in Table~\ref{table:all-search-spaces}. Their performance frontiers are shown in Fig.~\ref{fig:frontiers}.

\subsection{Experimental Setup}
\label{sec:exp_setup}
To ensure comparability we train all detection models with the same configuration and hyper-parameters. Ablation study results are reported on the $5$k COCO {\it val2017} dataset, whereas the final comparison is reported on the COCO {\it test-dev} dataset. 

{\bf Training setup}:
Training setup for COCO {\it val2017}: Each detection model is trained for $150$ epochs, or $277$k steps with a batch size of $64$ on COCO {\it train2017} dataset. Training is synchronized with $8$ replicas. Learning rate follows a step-wise procedure: it increases linearly from $0$ to $0.04$ in the first epoch then holds its value; The learning rate drops sharply to $0.1$ of its value at epoch $120$ and $140$, respectively. Gradient-norm clipping at $10$ was used to stabilize training.
In ablation studies, models that use MobileNetV2 as the backbone are warm-started from an ImageNet pre-trained checkpoint.

Training setup for COCO {\it test-dev}: Each model is trained for $100$k steps from scratch with a batch size of $1024$ over $32$ synchronized replicas with a cosine schedule for the learning rate~\cite{loshchilov2016sgdr}, which is decayed from $4$ to $0$. The schedule also comes with a linear warmup phase at the first $2$k steps. Following~\cite{sandler2018mobilenetv2, huang2017speed} to ensure comparability, we merged COCO {\it train2017} and {\it val2017} as training data.

All training and evaluation use $320\times 320$ input images. We do not employ drop-block or auto-augmentation or hyper-parameter tuning to avoid favoring a particular class of models in our comparison studies, and for fair comparison with some previous results in the literature.

{\bf Timing setup}:
All timing was performed on a Pixel 1 device with single-thread and a batch size of one using TensorflowLite's latency benchmarker\footnote{\url{https://www.tensorflow.org/lite/performance/benchmarks}}. Following the convention in MobileNetV2\cite{sandler2018mobilenetv2}, each detection model is converted into TensorflowLite flatbuffer format where the outputs are the box and class predictors immediately before non-max-suppression.

{\bf Architecture Search Setup}:
We follow the same controller setup as used in MNASNet~\cite{tan2019mnasnet}. The controller samples about $10$K child models, each taking $\sim 1$ hour of a TPUv2 device. To train a child model, we split COCO {\it train2017} randomly into a $111$k-{\it search-train} dataset and a $7$k-{\it search-val} dataset. We train for $20$ epochs with a batch size of $64$ on {\it search-train} and evaluate its mAP on {\it search-val}. Learning rate increases linearly from $0$ to $0.04$ in the first epoch, the follows a step-wise procedure that decays to $0.1$ of its value at epoch $16$. We used the same $320\times 320$ resolution for proxy task training to ensure that the estimated latency between the proxy task and the main task are identical. For the reward objective (Eq.~\ref{eq:reward}), we use $w=-0.3$, estimated from a few trial runs, for all search experiments.

After training, for \Mnasfpn we compute the performance frontier over all the sampled models, and fetch the top models at $166$ ms, $173$ ms and $180$ ms simulated latency. Then we increase the repeats from $3$ and $5$ to generate a  total of $3\times 3=9$ models. Among them we extract the performance frontier by only keeping models that are not dominated in both latency and mAP by any other model. 

\subsection{Discovered Architectures}
\label{sec:arch}


\begin{figure}[!t]
    \centering
    \includegraphics[width=0.47\textwidth]{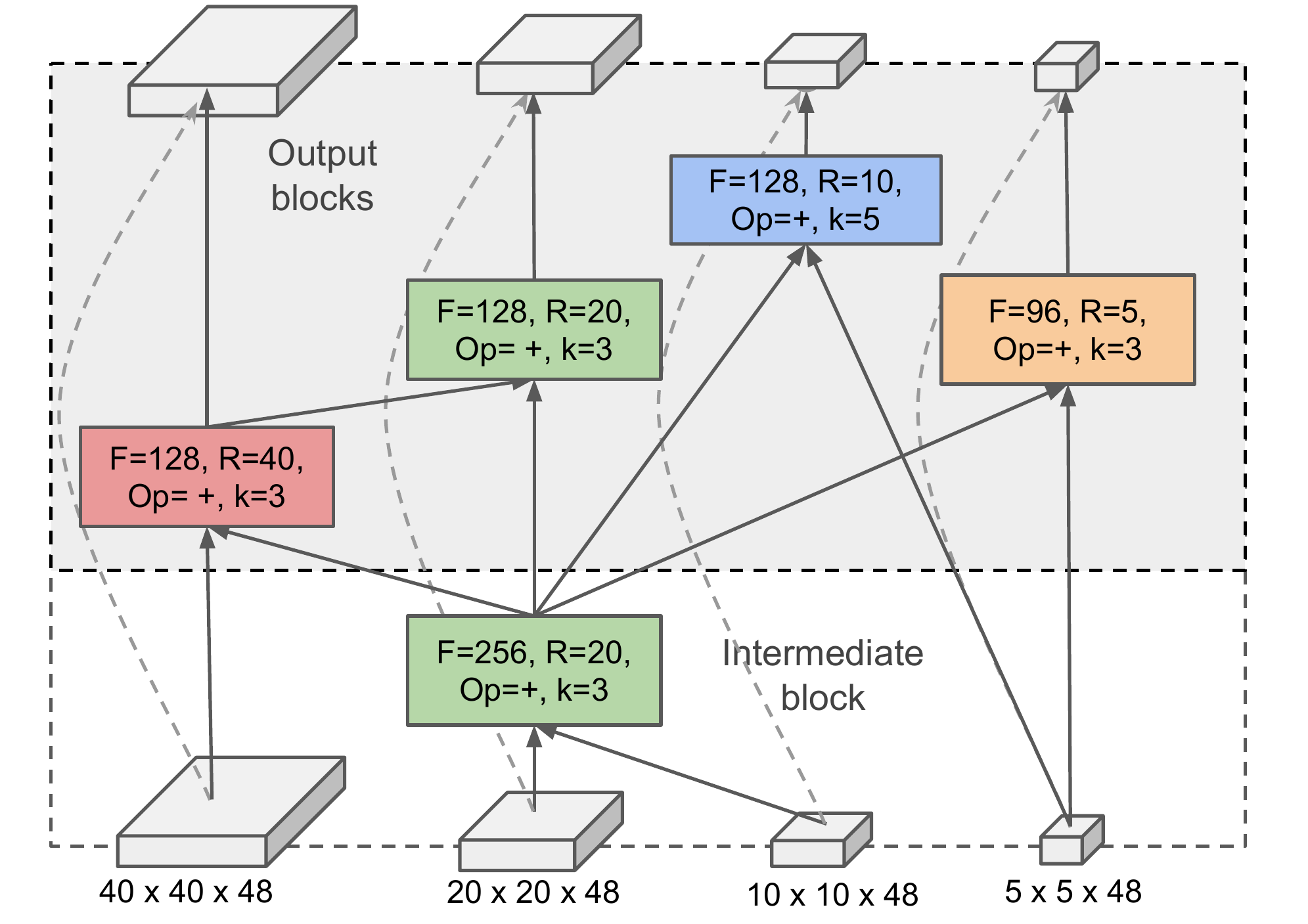}
    \caption{Visualization of a \Mnasfpn cell architecture found via latency-aware search. Both the inputs and outputs, represented as boxes with rounded edges, consist of four feature maps at $C_3$ to $C_6$, respectively. Each rectangle box represents a \Mnasfpn block whose internal structure is outlined in Fig.~\ref{fig:search_space}. The box also contains architectural parameters such as channel size $F$ and resolution $R$ for the intermediate feature, the merging operation $Op$, and the kernel size $k$ of the depthwise convolution. Finally, all outputs receive cell-wide residuals (dashed arrows) from the input with the corresponding resolution. Note that although the search allows for a maximum of $5$ intermediate blocks, only one was chosen.}
    \label{fig:arch}
\end{figure}

\begin{figure}[!t]
    \centering
    \includegraphics[width=0.47\textwidth]{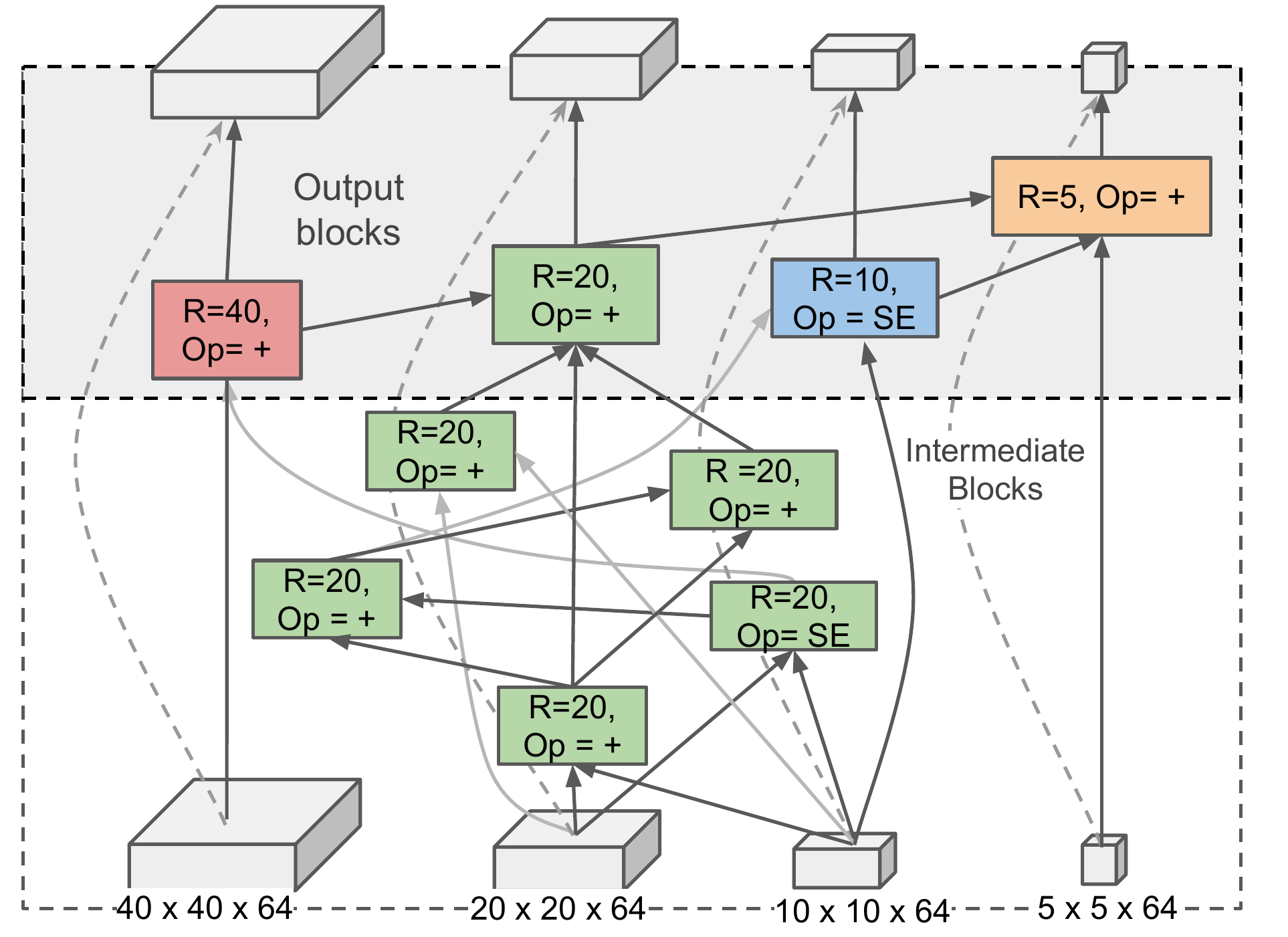}
    \caption{Visualization of a NAS-FPNLite-S cell architecture found via latency-aware search on the NAS-FPNLite search space. Each rectangle describes the resolution $R$ and merge operation (sum or SE) for the feature generation process. The channel sizes and kernel sizes are fixed to $64$ and $3$, respectively, according to NAS-FPNLite~\cite{ghiasi2019fpn}. }
    \label{fig:arch_nasfpn}
\end{figure}

We inspect a top-performing \Mnasfpn architecture in Fig.~\ref{fig:arch} and a NAS-FPNLite-S architecture in Fig.~\ref{fig:arch_nasfpn}. Both models have a similar latency as NAS-FPNLite. The comparison shows that:

    First, \Mnasfpn is the most compact. Despite both given $5$ internal blocks, \Mnasfpn only uses one block, whereas NAS-FPNLite-S uses $5$, and places all of them at the same resolution. \Mnasfpnnospace's compactness may be a product of 1) its ability to prune unused blocks and 2) the expansions in IRB that increases the capacity for each block.
    
    Second, the Squeeze-and-excite (SE) option to merge features is never used. This is an interesting discovery as SE was quite popular in the classification backbone.
    
    Third, both \Mnasfpn and NAS-FPNLite-S favor the $20\times20$ resolution for the intermediate features. This choice was also persistent among multiple search runs and multiple variations of search spaces.

\begin{figure}[!t]
    \centering
    \includegraphics[width=0.47\textwidth]{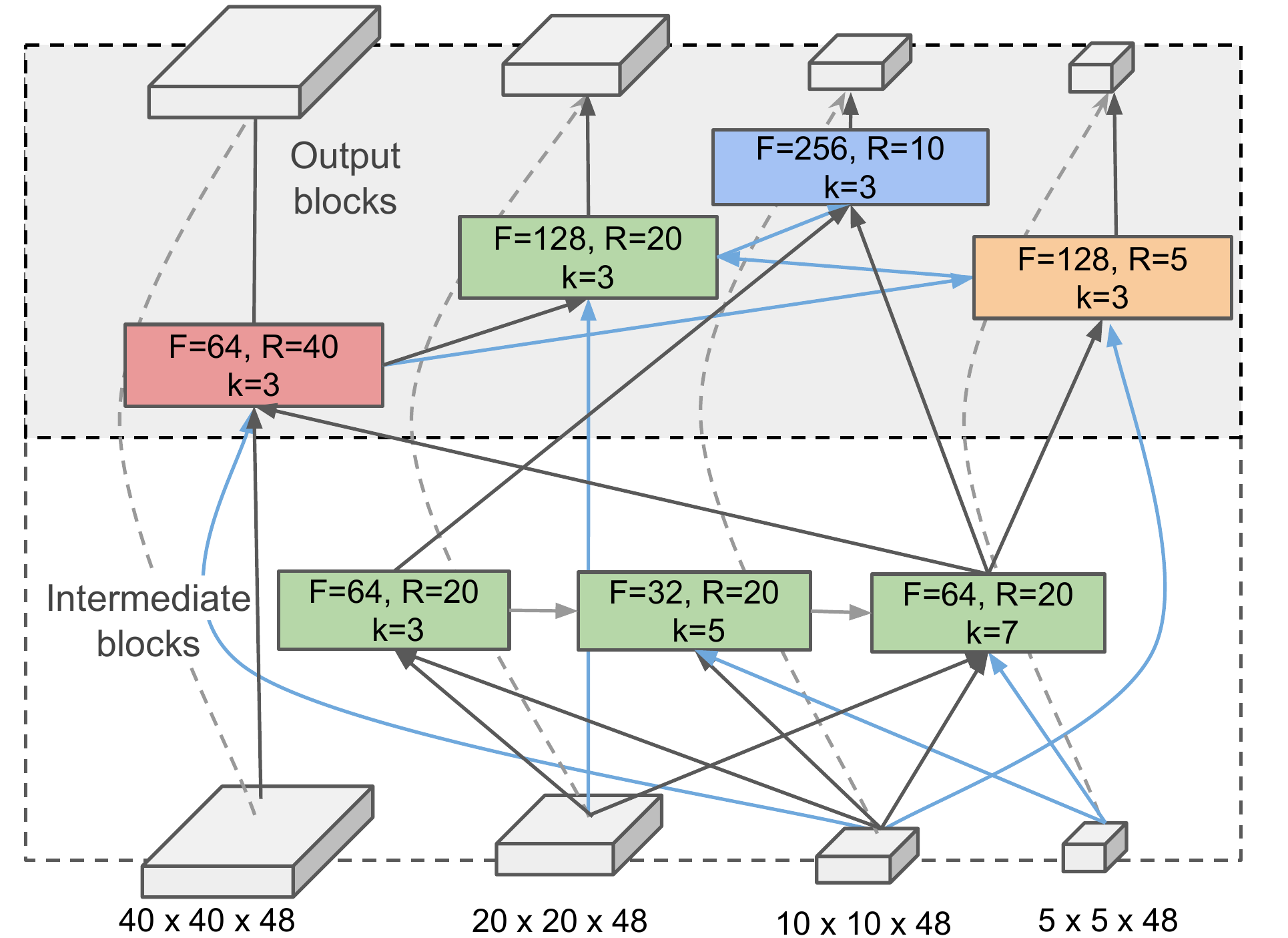}
    \caption{Visualization of a Conn-Search cell architecture with maximum in-degree $D=4$. Each rectangle describes the expansion size $F$, resolution $R$, and kernel size $k$ for the feature generation process. The merge operation is fixed to be summation. Blue arrows indicate the additional connections compared to \Mnasfpn in Fig.~\ref{fig:arch} where all intermediate blocks are treated as one agglomerate block. }
    \label{fig:connsearch_arch}
\end{figure}

Fig.~\ref{fig:connsearch_arch} shows a Conn-search architecture with $D=4$. 
First, similar to \Mnasfpnnospace, the resolutions of the intermediate features all concentrate around $20\times 20$. Second, almost in all cases only $2$ or $3$ features are merged. Therefore, either allowing $4$ input connections was already excessive, or the current search space is at the limit of what the search algorithm can handle. 

\subsection{Latency Breakdowns}
We divide a \Mnasfpn architecture into the feature extractor backbone, the detection head, and the ``predictor", which is a set of full convolutions followed by class predictors and box decoders. These full convolutions are $C\times C$ in size, where $C$ is the same parameter that describes the channel size of \Mnasfpn's outputs. Therefore, our search affects both the head and the predictor part of the network.

\begin{figure}[!t]
    \centering
    \includegraphics[width=0.47\textwidth]{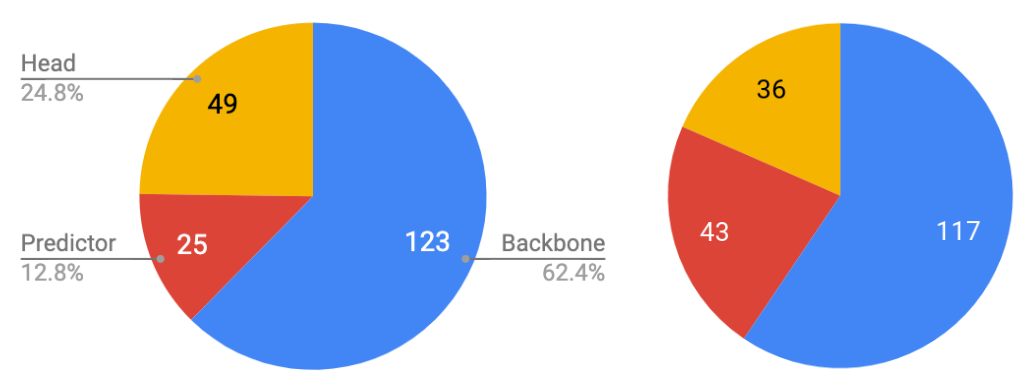}
    \caption{Latency breakdown of \Mnasfpn (left) and NAS-FPNLite (right). Both models have around $200$ms latency, out of which $40\%$ is reserved for the detection head as well as the box and class predictors, which we optimize in this paper. The \Mnasfpn model is $1.1$ mAP higher than the NAS-FPNLite model.}
    \label{fig:profile}
\end{figure}

To put the improvement on the \Mnasfpn detection head into perspective, we plot the latency breakdown of two $200$-ms models, namely \Mnasfpn with $5$ repeats ($25.5$ mAP) and NAS-FPNLite with $6$ repeats ($24.4$ mAP). 

As shown in Fig.~\ref{fig:profile}, our search affects around $80$ ms or $40\%$ of the total running time. 
\Mnasfpn ($C=48$) learns to allocate nearly $2\times$ more computational resources towards the head than the predictor, whereas NAS-FPNLite ($C=64$) allocates less resources towards the head than to the predictor. This suggests that more significance should be associated with early feature fusion in the detection head than with predictor capacity. 

The analysis above also indicates that as the detection head becomes more efficient with \Mnasfpn, the backbone, totaling around $60\%$ of run time, now becomes the performance bottleneck. Since joint search of backbone and head is outside the scope of this paper, it is reasonable to assess all improvements in the paper relative to the latency budget excluding the backbone. 

\subsection{Ablation on IRB}
\label{sec:ablation-irb}
To evaluate our primary contribution of re-introducing IRB into the detection head, we compare in Fig.~\ref{fig:comp} \Mnasfpn with NAS-FPNLite-S and No-Expand. 

\Mnasfpn and NAS-FPNLite-S share the use of latency-aware search and differ in the search space. We see that a \Mnasfpn at $187$ ms is more accurate than a NAS-FPNLite-S  model at $201$ ms, suggesting that the overall design of the \Mnasfpn search space contributes to almost all the improvements over NAS-FPNLite. 

\begin{figure}[!t]
    \centering
    \includegraphics[width=0.5\textwidth]{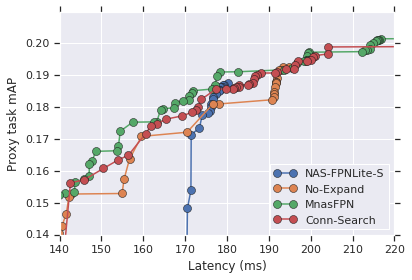}
    \caption{{\bf Proxy} task performance vs. {\bf simulated} latency frontiers of various search spaces. This figure represents the NAS controller's view on the problem, where latency is simulated using LUT and quality is computed on the proxy task, which correlates with but is not directly comparable to mAPs of the real task.}
    \label{fig:frontiers}
\end{figure}

\Mnasfpn and No-Expand differ only in the use of expansions in the \Mnasfpn block. No-Expand's performance is significantly below that of \Mnasfpnnospace. A closer inspection of the learned architectures shows that the model reduces the channel size $C$ to $16$ while increasing the number of intermediate nodes. This is a sub-optimal design strategy, on which the NAS controllers got stuck repeatedly. As a result, the entire performance frontier (during search) seems sub-optimal compared to those of other searches (Fig.~\ref{fig:frontiers}).

\subsection{Ablation on Latency-aware Search}
\label{sec:ablation-lat}
Our work is the first to introduce latency-aware training in architecture search for object detection. To investigate the gain of the latency signal, we compare \Mnasfpn with NAS-FPNLite and NAS-FPNLite-S. 

\begin{figure}[!t]
    \centering
    \includegraphics[width=0.47\textwidth]{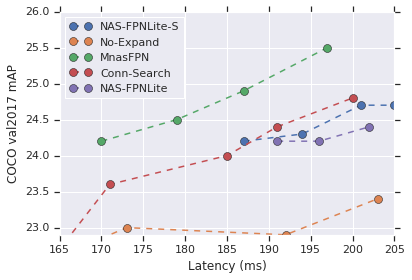}
    \caption{Performance comparisons between \Mnasfpn and various ablation designs. Latency is measured on Pixel 1 and mAP is computed on COCO {\it val2017}.}
    \label{fig:comp}
\end{figure}

According to Fig.~\ref{fig:comp}, \Mnasfpn shows a superior latency-accuracy tradeoff than NAS-FPNLite. At $187$ ms, \Mnasfpn achieves $24.9$ mAP that is unmatched even by the NAS-FPNLite model at $205$ ms. While the latency differential constitutes a mere $9\%$ in terms of end-to-end latency, it amounts to around $22\%$ improvement considering the latency portion excluding the backbone. 

NAS-FPNLite-S also performs better than NAS-FPNLite, but only by a moderate amount. This indicates that the MNASNet-styled latency-aware search is an effective strategy overall, but the primary factor of \Mnasfpnnospace's success is instead the search space design.

\subsection{Connectivity Search}
To assess whether the \Mnasfpn search space was sufficiently large, we compare with Conn-Search (Sec.~\ref{sec:conn}) where each block can take a maximum of $D=4$ inputs.



As shown in Fig.~\ref{fig:comp}, despite having a larger search space that subsumes \Mnasfpnnospace, Conn-Search has a suboptimal latency-accuracy tradeoff. In Fig.~\ref{fig:frontiers} we see that its performance frontier on the proxy task is slightly worse than that of \Mnasfpnnospace, suggesting that the controller is unable to sufficiently explore the search space. Table~\ref{table:all-search-spaces} shows that the cardinality of Conn-Search is roughly $10^{42}$, greatly surpasses the cardinalities of the two known successful applications of the MnasNet framework: MnasNet ($10^{13}$) and NAS-FPN ($10^{22}$).

This result reiterates the significance of the co-adaptation of search spaces and search algorithms. While it is tempting to believe that NAS eliminates the need for manual tuning, and that one only needs to innovate a sufficiently powerful search space that subsumes all search spaces, the reality is that the search algorithm is not yet powerful enough to address arbitrarily large search spaces. Therefore, iterative shrinking and co-adaptation of search spaces, as practiced in the original NAS paper~\cite{zoph2016neural}, are still relevant.

\subsection{Ablation on SDO}
\label{sec:ablation-sdo}
To understand the impact of SDO, we disable SDO of the \Mnasfpn architectures with $4$ and $5$ repeats, respectively. Models with no SDO will perform $1\times1$ convolution before resizing, which will be less economical for down-sampling operations, and the discovered \Mnasfpn architecture as it is dominated by down-sampling operations (Fig.~\ref{fig:arch}).

Unsurprisingly, we see from Table~\ref{table:sdo-ablation} that disabling SDO does not affect the mAP, but would lead to a $8$ to $11$ms ($4-6\%$) latency regression. Similarly if we consider the portion of the network without the backbone, this amounts to $12\%$ to $14\%$ of the latency that is ``optimizable``. Given this strict dominance we conclude that the effectiveness of SDO is sufficiently evident and do not conduct search experiments without SDO for the ablation study.

\begin{table}[t]
\centering
\vspace{0pt}
\scalebox{0.8}{
\begin{tabular}{c|c|c|c|c|c}
\toprule[0.2em]
Model & Repeats & mAP & Latency (ms) & MAdds (B) & Params (M)\\
\toprule[0.2em]
\Mnasfpn & $4$ & $24.9$ & $187$  & $0.90$ & $2.5$ \\	
	& $5$ & $25.5$ & $196$ & $0.94$ & $2.6$ \\	
\toprule[0.1em]
No SDO & $4$ & $24.9$ & $195$  & $0.94$ & $2.5$ \\
& $5$ & $25.5$ & $207$ & $1.0$ & $2.6$ \\
\toprule[0.2em]
\end{tabular}
}
\caption{Ablation study of SDO. SDO does not affects parameters that much but reduces both MAdds and latency.}
\label{table:sdo-ablation}
\end{table}

\subsection{Performance Comparison on COCO Test-dev}
\label{sec:generalizability}
\label{sec:testdev}

We compare \Mnasfpn under different backbones and with other state-of-the-art on-device detection heads. \footnote{The COCO {\it test-dev} mAPs are produced from a more time-consuming setup (Sec.~\ref{sec:exp_setup}) than the COCO {\it val2017} mAPs for internal comparisons, hence the performances differ slightly as well.} 






As shown in Table~\ref{tab:testdev}, with the same MobileNetV2 backbone, \Mnasfpn achieves {\bf $\mathbf{1.0}$ mAP improvement over NAS-FPNLite}. Furthermore, \Mnasfpn is $10\%$ faster in end-to-end latency, or {\bf $\mathbf{25\%}$ faster} in terms of latency incurred outside the backbone. 

Since SSDLite is generally much faster than \Mnasfpn, we compare the two either by applying width-multipler or changing the backbone. With a $0.7$ width-multiplier on both head and backbone, \Mnasfpn with MobileNetV2 achieves {\bf $\mathbf{1.8}$ higher mAP compared with SSDLite} with MobileNetV3 at around $120$ ms. Here the MobileNetV3 results use the channel-halving trick, which tends to reduce latency with no mAP degradation, while our results do not. Removing this trick for both shows a further $20$ ms latency advantage for \Mnasfpnnospace. 

When paired with MobileNetV3 backbone, \Mnasfpn is $3.4$ mAP higher than SSDLite with MobileNetV2 at around $165$ ms. It is both faster and $2.5$ mAP higher than SSDLite with MnasNet-A1 backbone. 

Therefore, we conclude that \Mnasfpn compares favorably to both SSDLite and NAS-FPNLite head in its ability to trade off latency with accuracy.

%% file: conclusions.tex
\section{Conclusion}
In this paper, we show the benefits of treating object detection as a first-class citizen in NAS. Unlike previous work that transfers learned backbone from classification, our work directly searches for object detection architectures. Additionally, we design the search process and, more importantly, the search space to incorporate knowledge about the targeted platform. Our proposed \Mnasfpn search space has two innovations. First, \Mnasfpn incorporates inverted residual blocks into the detection head, which is proven to be favored on mobile CPUs. Second, \Mnasfpn restructured the reshaping and convolution operations in the head to facilitate efficient merging of information across scales. 

Through detailed ablation studies, we've discovered that both innovations in the search space are necessary for the performance boost. On the other hand, further expanding the search space in feature map connectivity seems to overwhelm the NAS framework. As a result, we conclude that the proposed \Mnasfpn search space may be close to the capacity {\it of this controller}. As the controller becomes more powerful, the \Mnasfpn~with connectivity search could become viable again. 

On COCO {\it test-dev} \Mnasfpn leads to a $25\%$ improvement in non-backbone latency over NAS-FPNLite. The improvements are so substantial that the rest of the network becomes the bottleneck for performance improvements. For example, the backbone, which currently occupies over $60\%$ of the total latency, could be searched either conditioning on or jointly with the \Mnasfpn head. This seems promising with our anecdotal evidence in Table~\ref{tab:testdev} that \Mnasfpn pairs well with MobileNetV3 and depth-multiplied MobileNetV2 backbones. 
While the cardinality of a joint-search of backbone and the head is challenging for our current controller, recent one-shot NAS methods are opening avenues for more ambitious search spaces, of which \Mnasfpn could be an ideal component.

%% file: appendix.tex
\section{Appendix}
\subsection{Search space cardinality comparison}
\label{sec:search_space_size}
{\bf NAS-FPNLite-S}: There are $9$ nodes in total, where the $i$-th node has two choices for the combine operation, and $choose(i+4, 2)$ choices for picking a pair of inputs. The first $5$ are internal nodes, and each have $4$ resolutions choices. The last $4$ are output nodes, whose orders are permuted with $permute(4)$ possibilities. This gives a total search space size of:
\begin{align*}
    2^9 4^5 permute(4) \prod_{i=0}^8 choose(i+4, 2) \approx 2\times 10^{22}
\end{align*}

{\bf No-Expand}: In addition to the NAS-FPNLite-S search space, No-Expand additionally grants $3$ kernel sizes for each node. It also have $6$ choices for the globally-shared channel size $C$, giving a total search space size of:
\begin{align*}
    2\times 10^{22} \times 3^{9} \times 6 \approx 2.4 \times 10^{27}
\end{align*}

{\bf \Mnasfpn}: In addition to the No-Expand search space, \Mnasfpn additionally searches for channel sizes for the merged features for all $9$ nodes, each with $7$ choices. The total search space size is:
\begin{align*}
    2\times 10^{27} \times 7^9 \approx 10^{31}
\end{align*}

{\bf Conn-Search}: Finally, connectivity search allows for $choose(i+4, 4)$ choices for each node, which is $(i+2)(i+1) \times$ more possibilities than that in \Mnasfpnnospace. It does not search for combine operations, so each node has $2\times$ fewer choices. Therefore the total search space size is:

\begin{align*}
    10^{31} \prod_{i=0}^8 (i+2)(i+1) / 2^{9} \approx 3 \times 10^{42}
\end{align*}